\documentclass[letterpaper, 10 pt, journal, twoside]{IEEEtran}
\usepackage[dvipsnames]{xcolor}

\usepackage{todonotes}

\usepackage{amsmath,amsfonts,amssymb}
\usepackage{algorithmic,algorithm}
\usepackage[export]{adjustbox} % also loads graphicx
\usepackage[noadjust]{cite}

\usepackage{lineno}
\usepackage{soul,color}
\usepackage{comment}

\usepackage{multirow, hhline}
\usepackage{subfloat}
\usepackage{mathtools, mathdots}
\usepackage{hyperref}
\hypersetup{
    colorlinks=true,
    linkcolor=blue,
    filecolor=magenta,      
    urlcolor=blue,
}
\usepackage{caption, subcaption} 
\usepackage{textcomp}
\usepackage{gensymb}
\usepackage{diagbox}
\usepackage{enumitem}
\usepackage{siunitx}
\usepackage{lipsum}
\usepackage{optidef}
\usepackage{eso-pic}

\usepackage{caption}
\usepackage{makecell}

\DeclareMathAlphabet\mathbfcal{OMS}{cmsy}{b}{n}

\newcommand{\smallsim}{\smallsym{\mathrel}{\sim}}
\makeatletter
\newcommand{\smallsym}[2]{#1{\mathpalette\make@small@sym{#2}}}
\newcommand{\make@small@sym}[2]{%
  \vcenter{\hbox{$\m@th\downgrade@style#1#2$}}%
}
\newcommand{\downgrade@style}[1]{%
  \ifx#1\displaystyle\scriptstyle\else
    \ifx#1\textstyle\scriptstyle\else
      \scriptscriptstyle
  \fi\fi
}
\makeatother

\begin{document}
\AddToShipoutPictureBG*{%
  \AtPageUpperLeft{%
    \setlength\unitlength{1in}%
    \hspace*{\dimexpr0.5\paperwidth\relax}%%  change \dimexpr0.5\paperwidth\relax appropriately
    \makebox(0,-0.75)[c]{\parbox{0.8\textwidth}{\centering \small This paper has been accepted for publication in the IEEE Robotics and Automation Letters. Please cite as: Montano-Oliv\'an, L., Placed, J. A., Montano, L., and L\'azaro, M. T. (2024), G-Loc: Tightly-coupled Graph Localization with Prior Topo-metric Information. IEEE Robotics and Automation Letters (RA-L). DOI: 10.1109/LRA.2024.3457383.}}%
}}

\title{G-Loc: Tightly-coupled Graph Localization with Prior Topo-metric Information}

\author{Lorenzo~Montano-Oliv\'an,
        Julio~A.~Placed,
        Luis~Montano,
        and~Mar\'ia~T.~L\'azaro, \textit{Member, IEEE}% <-this % stops a space
\thanks{Manuscript received: May 6, 2024; Revised: June 27, 2024; Accepted: August 25, 2024.}% <-this % stops a space
\thanks{This paper was recommended for publication by Editor Sven Behnke upon evaluation of the Associate Editor and Reviewers' comments. This work was partially supported by DGA\_FSE T73\_23R and PTAS-20211016 DIGIZITY. }% <-this % stops a space
\thanks{Lorenzo~Montano-Oliv\'an, Julio A.~Placed, and Mar\'ia T.~L\'azaro are with the Instituto Tecnol\'ogico de Arag\'on (ITA), Mar\'ia de Luna 7, 50018 Zaragoza, Spain {\tt\small \{lmontano, jplaced, mtlazaro\}@ita.es}}% <-this % stops a space
\thanks{Luis Montano is with the Instituto de Investigaci\'on en Ingenier\'ia de Arag\'on (I3A), Universidad de Zaragoza, Mar\'ia de Luna 1, 50018 Zaragoza, Spain {\tt\small montano@unizar.es}}% <-this % stops a space
\thanks{Digital Object Identifier (DOI): see top of this page.}% <-this % stops a space
}

% \markboth{IEEE ROBOTICS AND AUTOMATION LETTERS. PREPRINT VERSION. ACCEPTED
% AUGUST, 2024}
% {Montano-Oliv\'an \MakeLowercase{\textit{et al.}}: G-Loc: Tightly-coupled Graph Localization} 

\maketitle

\begin{abstract}
    Localization in already mapped environments is a critical component in many robotics and automotive applications, where previously acquired information can be exploited along with sensor fusion to provide robust and accurate localization estimates.
    In this work, we offer a new perspective on map-based localization by reusing prior topological and metric information. Thus, we reformulate this long-studied problem to go beyond the mere use of metric maps.
    Our framework seamlessly integrates LiDAR, iner\-tial and GNSS measurements, and cloud-to-map registrations in a sliding window graph fashion, which allows to accommodate the uncertainty of each observation.
    The modularity of our framework allows it to work with different sensor configurations (\textit{e.g.,} LiDAR resolutions, GNSS denial) and environmental conditions (\textit{e.g.,} mapless regions, large environments). 
    We have conducted several
    validation experiments, including the deployment in a real-world automotive application, demonstrating the accuracy, efficiency, and versatility of our system in online localization.
\end{abstract}

\begin{IEEEkeywords}
Robot Localization, Graph Optimization, Topo-metric Map.
\end{IEEEkeywords}

\IEEEpeerreviewmaketitle

%%%%%%%%%%%%%%%%%%%%%%%%%%%%%%%%%%%%%%%%%%%%%%%%%%%%%%%%%%%%%%%%%%%%%%%%%%%%%%%%
%%%%%%%%%%%%%%%%%%%%%%%%%%%%%%%%%%%%%%%%%%%%%%%%%%%%%%%%%%%%%%%%%%%%%%%%%%%%%%%%
\section{Introduction}
\IEEEPARstart{R}{obust} and precise (global) localization is essential for autonomous robots and self-driving vehicles. Simultaneous Localization and Mapping (SLAM) solutions~\cite{cadena16, campos21} have emerged as a response to 3D reconstruction and online localization, growing exponentially in recent years also in the autonomous driving realm~\cite{bresson17}.
However, there are two situations where these methods suffer from limitations.
On the one hand, they assume that the robot is teleoperated, or that the motion actions are known \emph{a priori}. Active SLAM tackles this by integrating online path planning into the problem~\cite{placed23}.
On the other hand, it is desirable for many robotic applications to navigate repeatedly in previously visited (and mapped) areas without the need of further updating the built model of the environment. For example, autonomous vehicles often navigate familiar routes, and robots typically operate in known environments when not engaged in exploratory tasks.
Using a previously built environment model in such situations can scale down the problem of SLAM to map-based localization.
This work specifically focuses on the latter, aiming to leverage existing \emph{prior} information to address localization.

\begin{figure} [t!]
    \centering
    \includegraphics[width=0.9\linewidth]{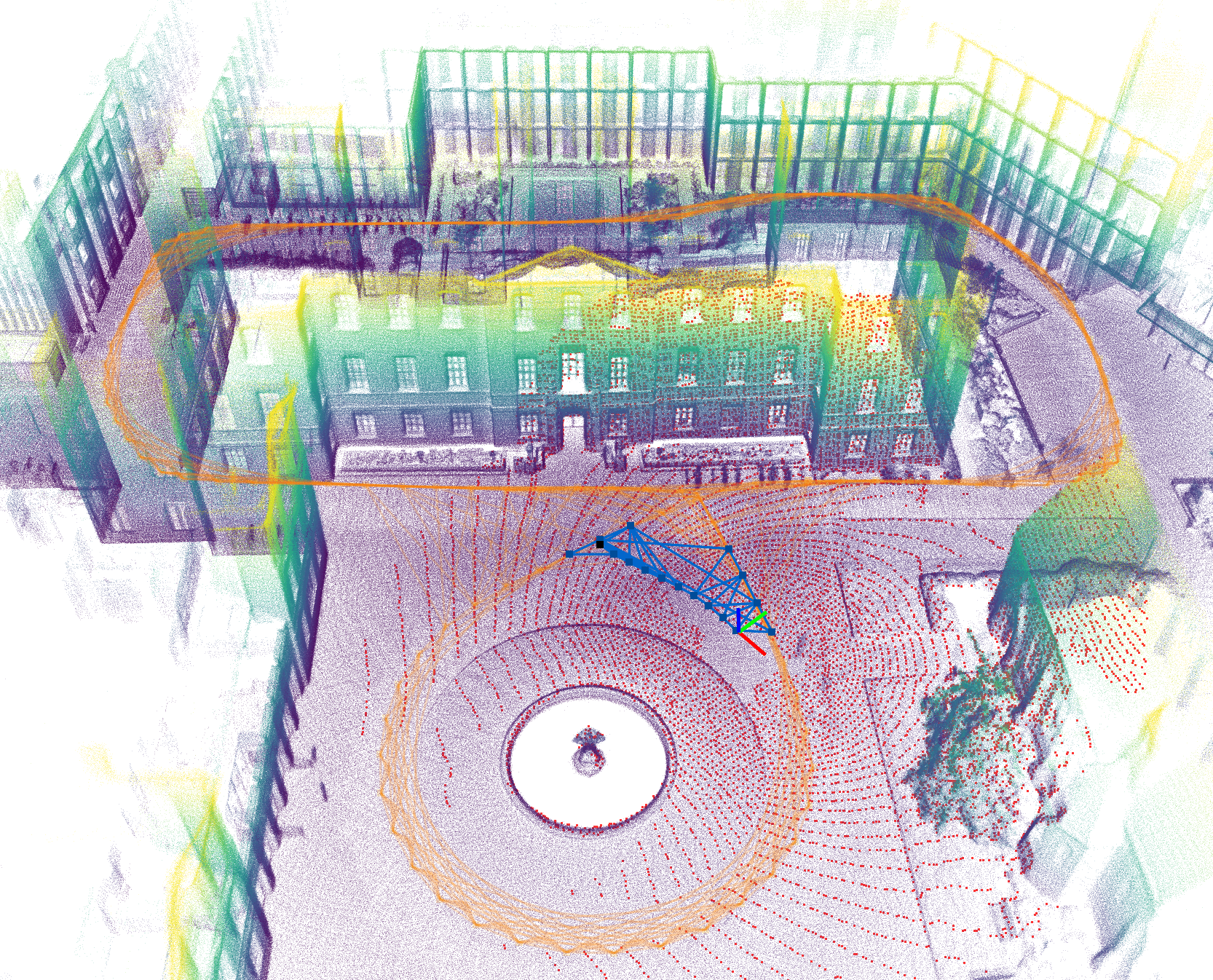}
    \setlength{\belowcaptionskip}{-8pt}
    \caption{Visualization of G-Loc in a sequence of the \emph{Newer College Extension} dataset~\cite{zhang21}. The prior topological (orange pose-graph) and metric (colored pointcloud) model is exploited for online robust localization. An \emph{active graph} (blue) is optimized, containing certain vertices from the prior graph and the most recent robot states.}
    \label{fig:gloc_general}
\end{figure}

At first glance, without the need of updating an environment model, the use of Global Navigation Satellite System (GNSS) in \mbox{(semi-)}urban areas may seem convenient for localization, although its use alone has long been proven to be neither robust nor accurate enough for the required applications~\cite{levinson07}. 
Combining measurements from multiple sensors and finding correspondences with a reference map model is key to robust, reliable and accurate localization in automotive applications.
Most of the existing work tackles this task by using filters for measurement integration, and by exploiting dense high-resolution point cloud maps previously generated by SLAM algorithms~\cite{yoneda14, kato15, nagy18, shan23}. 
However, modern SLAM systems, based on probability and graph theories~\cite{cadena16}, contain much more information beyond the resulting reconstruction: a rich underlying graph that captures the localization estimates and accommodates the observation uncertainties.
This valuable topological information has been overlooked in the literature.

In this work, we present G-Loc (see Figure~\ref{fig:gloc_general}), a novel tightly-coupled graph localization approach that exploits prior topo-metric knowledge of the environment. We propose to go beyond purely geometric maps and use all the information provided by modern SLAM systems. In addition, we efficiently fuse this information with LiDAR (Light Detection And Ranging), IMU (Inertial Measurement Unit), and GNSS observations to create a robust and accurate localization system.

The rest of the paper is organized as follows. Section~\ref{S:2} discusses related work and the contributions of this paper, Section~\ref{S:3} formally defines the problem, Section~\ref{S:4} details the components of our system, and Section~\ref{S:5} contains the experimental results. This letter is concluded in Section~\ref{S:6}.

%%%%%%%%%%%%%%%%%%%%%%%%%%%%%%%%%%%%%%%%%%%%%%%%%%%%%%%%%%%%%%%%%%%%%%%%%%%%%%%%
%%%%%%%%%%%%%%%%%%%%%%%%%%%%%%%%%%%%%%%%%%%%%%%%%%%%%%%%%%%%%%%%%%%%%%%%%%%%%%%%

\section{Related Work}\label{S:2}

Levinson \textit{et al.}~\cite{levinson07} are among the first to exploit a previously-built map for localization. They correlate LiDAR intensity measurements with the prior map (a grid representation containing a Gaussian distribution of intensity in each cell) using a particle filter. 
They also incorporate GNSS measurements to limit particle dispersion. However, this and other related work~\cite{wolcott17},
%levinson10
reduce the estimation problem to 3 degrees of freedom, assuming that the rest are estimated, \textit{e.g.,}~from an IMU.
\cite{wan18} estimates a global 3D pose instead after incorporating altitude into the distribution of the cells. 

A different approach is followed in~\cite{poggenhans18}, where the reference map consists of road primitives. An Unscented Kalman Filter (UKF) is used to combine GNSS, odometry and the matching between the prior map and the online primitives extracted from images via convolutional neural networks. 
The challenge of maintaining and matching a dense point cloud map is also addressed in~\cite{javanmardi19}, where such representations are transformed into a sparse multi-layer vector map, and~\cite{SecoSensors22}, where features are extracted from lateral galleries in tunnel-like environments.

More recently, advances in point cloud registration have brought these methods to the forefront of map-based localization, whether feature-based or not.
Yoneda and Mita~\cite{yoneda14} first matched LiDAR measurements to a georeferenced point cloud map using Iterative Closest Point (ICP)~\cite{besl92}. The limitations of this single sensor approach and the sensitivity of ICP to the initial guess are addressed in subsequent works, all of them being filter-based~\cite{kato15, nagy18, ma19, jeong20, shan23}.
Nagy and Benedek~\cite{nagy18} fuse point cloud and semantic matching with GNSS measurements to localize a robot in a high resolution 3D semantic cloud.
In~\cite{wolcott14}, instead of matching the online LiDAR data to the reference point cloud map, the latter is rendered in 2D; this allows to match monocular images.
In~\cite{wu21} LiDAR Odometry (LO) and IMU data are combined in a UKF to improve the matching of the current scan with a portion of the prior point cloud map; this portion is given by cropping the map around the GNSS pose estimation.
Autoware~\cite{kato15}, a widely used open-source library constantly under development, features global localization in a prior map (either in the form of a point cloud or a high-definition vector map). This approach relies on aligning the input clouds with the reference map using the Normal Distribution Transform (NDT)~\cite{biber03} technique. The optimized transformation is integrated with IMU and GNSS measurements within an extended Kalman filter module.
Xia \textit{et al.}~\cite{xia23} combine GNSS, IMU and NDT-based map matching on a Kalman filter. However, they use a light map instead of selecting regions of interest from a dense map. On the downside, this requires a special offline process.
\textit{PoseMap}~\cite{egger18} incorporates the trajectory into the prior model.
Each robot pose is assigned a submap of surfels and the online cloud is aligned only with the $k$-closest neighbors, reducing the computational cost and allowing to use the full LiDAR range.

In contrast to the filter-based techniques mentioned above, a handful of methods have recently started to use a graph formulation, borrowing from the advances in graph SLAM and due to its improved robustness and accuracy~\cite{grisetti10}.
Dub\'e \textit{et al.}~\cite{dube20} estimate the robot's pose in a prior global map by extracting segments from a 3D point cloud and matching neighbors in the feature space. The resulting transformations are later fed into a pose-graph SLAM framework along with pre-integrated IMU measurements. 
\cite{guo21} also employs a pose-graph formulation for localization, combining wheel odometry, GNSS, and semantic matching with the reference map. 
In parallel to our work, Koide \textit{et al.}~\cite{koide24} recently proposed a map-based localization approach that tightly couples range and IMU data within a sliding window factor graph optimization.

The localization method proposed in this letter falls into the graph-based category.
G-Loc is a modular and optimized framework that adapts to a wide variety of sensor and environment configurations, capable of handling large-scale environments, GNSS denial, mapless regions, etc.
Unlike previous work~\cite{dube20,guo21,koide24} we go beyond geometric representations and fully exploit the knowledge gathered during a previous mapping task, somewhat following~\cite{egger18} but leveraging the pose-graph topology instead of just the trajectory poses.
More specifically, our main contributions are as follows: 
\begin{itemize}
    \item A novel approach to map-based localization that exploits the topology of the prior model both for an efficient search of regions of interest, and for accommodating their uncertainty in a graph optimization process. 
    \item A complete localization system that combines precise LiDAR-inertial odometry, GNSS and map matching constraints. It can accurately locate a robot under various sensor configurations and environmental challenges.
    \item Thorough experiments demonstrating the accuracy and versatility of our system compared to state-of-the-art localization and SLAM methods, as well as real world deployment in an autonomous driving scenario.
\end{itemize}

%%%%%%%%%%%%%%%%%%%%%%%%%%%%%%%%%%%%%%%%%%%%%%%%%%%%%%%%%%%%%%%%%%%%%%%%%%%%%%%%
%%%%%%%%%%%%%%%%%%%%%%%%%%%%%%%%%%%%%%%%%%%%%%%%%%%%%%%%%%%%%%%%%%%%%%%%%%%%%%%%
\section{Problem Formulation}\label{S:3}

Map-based localization aims to determine the pose of a robot or sensor within an existing map of the environment. This can intuitively be seen as a coordinate transformation process, which consists of establishing correspondences between the map information and the current perception~\cite{thrun2002probabilistic}.

First of all, let us denote the map and robot base frames by {\small$m$} and {\small$r$}, respectively. To simplify the notation, all the sensors will be assumed to be in {\small$r$}.
In its most basic form, map-based 3D localization can be formulated as the problem of estimating the robot poses that best match the sensor observations with each other and with the map, \textit{i.e.,}~finding the \textit{maximum a posteriori} estimates
\begin{equation}
       \mathcal{T}^\star 
       \triangleq \underset{\mathcal{T}}{\mathrm{argmax}} \ 
       p(\mathcal{T} \, | \, \mathcal{M}, \mathcal{Z})\,, \label{eq:map_estimate} 
\end{equation}
where {\small$p(\cdot)$} denotes probability, {\small$\mathcal{M}$} is the prior map, {\small$\mathcal{Z}$} is the set of sensor measurements, and {\small$\mathcal{T} \triangleq \{{}^m\boldsymbol{T}_{r_i} \in SE(3) \, | \, i\in\mathcal{K}_t\}$} is the set of relative transformations between the map and the latest {\small$N_\mathcal{K}\texttt{+}1$} robot frames. To maintain tractability, the set of frame indices {\small$\mathcal{K}_t\triangleq\{\max(0,t\texttt{-}N_\mathcal{K}),\dots,t\}$} has been limited in a sliding window fashion.

On the one hand, we assume that {\small$\mathcal{M}$} has been previously built by a graph-SLAM algorithm, and thus contains metric and topological information (in the form of point clouds and a pose-graph, respectively). Furthermore, the metric part is associated with the nodes of the graph in such a way that each vertex is assigned a local map (or \emph{submap}, {\small{$\mathcal{S}'$}}) of its surroundings (see upper part of Figure~\ref{fig:active_ref_graph}).
Then:
\begin{subequations}
    \begin{align}
    \mathcal{M} \triangleq \left\{ \mathcal{G} \,, \mathcal{S}_\ell \, | \, \ell \in\mathcal{R} \right\} \,, \\
   \mathcal{G} \triangleq \{ \mathcal{V}, \mathcal{E} \} \,,
    \end{align}
\end{subequations}
\noindent where {\small$\mathcal{R}\triangleq \{0, \dots, \dim(\mathcal{V})\}$} is the reference frame index set, and {\small$(\mathcal{V},\mathcal{E})$} are the sets of vertices and edges in the pose-graph. 
The $\ell$-th submap will be formed by concatenating the point clouds between the {\small$\ell$} and {\small$\ell-N_s$} LiDAR frames, as described in~\cite{montano24}, or just a region around the reference vertex. The latter can be easily obtained by splitting the resulting map of a typical SLAM process~\cite{shan2020lio} into grids.
Notice that the submaps of nearby vertices might overlap and that elements of the pose-graph will live in $SE(3)$. Additionally, the graph may contain GNSS information, making also the map georeferenced and aligned to the East, North, Up (ENU) global coordinate system.
On the other hand, the set of measurements collected up to time {\small$t$} consists of LiDAR point clouds, IMU observations between consecutive frames, and GNSS data: {\small $\mathcal{Z}_t \triangleq \{ \mathcal{Z}^{\text{LiDAR}}_i, \mathcal{Z}^{\text{IMU}}_{i,j}, \mathcal{Z}^{\text{GNSS}}_i \}_{i,j\in\mathcal{K}_t}$}.

Assuming (i) that the robot poses are normally distributed (\textit{i.e.,}~{\small$\boldsymbol{T}\triangleq \boldsymbol{\bar{T}} \exp(\boldsymbol{\delta})\in SE(3)$}, with {\small$\boldsymbol{\bar{T}}$} a large mean transformation and {\small$\boldsymbol{\delta}\in\mathbb{R}^6$} a random vector normally distributed around zero) and pairwise independent, (ii) the measurements are conditionally independent w.r.t. the poses, and (iii) LiDAR and IMU observations are combined to provide odometry estimates,~\eqref{eq:map_estimate} can be rewritten as follows:
\begin{gather}
    \mathcal{T}^\star = 
    \underset{\mathcal{T}}{\mathrm{argmin}} \ \boldsymbol{F}(\mathcal{T})        \label{eq:optimization} \\
    \nonumber\text{s.t.} \ \boldsymbol{F}(\mathcal{T}) =        
        \textstyle \sum\limits_{i,j\in\mathcal{K}_t} \| \boldsymbol{e}_{i,j}^{\text{LIO}}\|^2
        + \sum\limits_{i\in\mathcal{K}_t} \| \boldsymbol{e}_{i}^{ \mathcal{M}}\|^2
        + \sum\limits_{i\in\mathcal{K}_t} \| \boldsymbol{e}_{i}^{\text{GNSS}}\|^2 
        \,,
\end{gather}
\noindent where 
{\small$\|\boldsymbol{e}\|^2\triangleq (\boldsymbol{e}^T  \boldsymbol{\Sigma}^{\texttt{-}1}  \boldsymbol{e})\in\mathbb{R}$} is the quadratic error for a measurement with covariance {\small$\boldsymbol{\Sigma}$}. {\small$\boldsymbol{e}_{i,j}^{\text{LIO}}$} are the residual errors associated with the LiDAR-inertial odometry (LIO) estimates,
{\small$\boldsymbol{e}_{i}^{\mathcal{M}}$} are those associated with the correspondences between LiDAR observations and the prior map, and {\small$\boldsymbol{e}_i^{\text{GNSS}}$} are the translation errors associated with the GNSS measurements. 

%%%%%%%%%%%%%%%%%%%%%%%%%%%%%%%%%%%%%%%%%%%%%%%%%%%%%%%%%%%%%%%%%%%%%%%%%%%%%%%%
%%%%%%%%%%%%%%%%%%%%%%%%%%%%%%%%%%%%%%%%%%%%%%%%%%%%%%%%%%%%%%%%%%%%%%%%%%%%%%%%
\section{Methods}\label{S:4}

\begin{figure} [t!]
    \centering
    \includegraphics[width=0.98\linewidth]{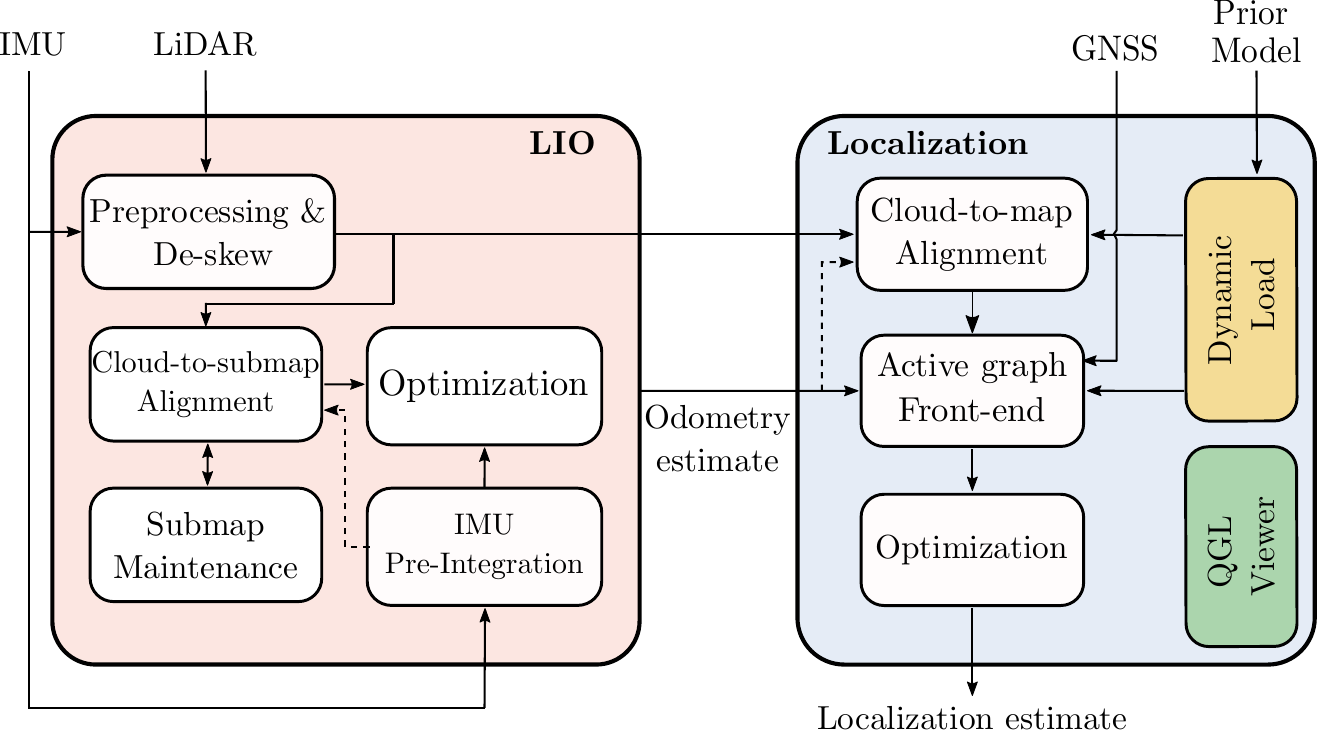}
    \caption{Block diagram of G-Loc. The dashed lines represent the initial guess fed  to the registration methods.}
    \label{fig:block_diag}
\end{figure}

G-Loc is divided into three main threads: LIO, Localization and Dynamic Loading, to which the following subsections are dedicated.
Figure~\ref{fig:block_diag} provides an overview of the system, showing the different modules and their connections. 
Thanks to its parallelized architecture, the entire system runs in real time within a ROS2 Humble~\cite{macenski2022robot} framework. It is able to use GPU-accelerated cloud registration 
and is robust to a wide range of sensor configurations with minimal parameter tuning.

LIO combines range and inertial measurements within a graph optimization framework. Online LiDAR data is aligned with most recent point clouds (which are contained in a submap), and high-frequency IMU data is pre-integrated for an efficient optimization.
The output of LIO is then fed into the localization module, along with GNSS measurements and the results from cloud-to-map registration.
A second pose-graph is optimized taking into account the previous constraints. Splitting the optimization problem into two different graphs improves the stability and convergence of the system and facilitates the noise configuration~\cite{shan2020lio}.
The dynamic loader allows to work with large prior maps by limiting the number of point clouds loaded in memory.
Finally, visualization is based on OpenGL and handled in a separate thread.

%%%%%%%%%%%%%%%%%%%%%%%%%%%%%%%%%%%%%%%%%%%%%%%%%%%%%%%%%%%%%%%%%%%%%%%%%%%%%%%%
\subsection{LiDAR-inertial Odometry (LIO)}\label{S:4a}

Relative measurements between consecutive robot poses are computed in a graph-based LIO framework. The incremental inertial state can be defined as
\begin{subequations}
    \begin{align}
    \mathcal{X} \triangleq \{ \boldsymbol{x}_i \,|\, i \in \mathcal{I}_t\} \,, \\
    \boldsymbol{x}_i \triangleq \{ \boldsymbol{T}_i, \ \boldsymbol{v}_i, \ \boldsymbol{b}_i \} \,,
    \label{eq:lio_state}
    \end{align}
\end{subequations}
\noindent where {\small$\boldsymbol{v}\in\mathbb{R}^3$} is the linear velocity and {\small$\boldsymbol{b}=(\boldsymbol{b}^g, \boldsymbol{b}^a)\in\mathbb{R}^6$}, with {\small$\boldsymbol{b}^g$}, {\small$\boldsymbol{b}^a$} in {\small$\mathbb{R}^3$} the gyroscope and accelerometer bias, respectively. The dimension of {\small$\mathcal{X}$} is constrained by {\small$\mathcal{I}_t\triangleq\{\max(0,t\texttt{-}N_\mathcal{I}),\dots,t\}$}, similarly to {\small$\mathcal{K}_t$}.
The map frame superindex notation has been dropped for readability.

First, a preprocessing step is performed. The input cloud is downsampled to a desired voxel resolution, and points too close/distant are removed. 
Then, a \emph{de-skew} operation is performed to mitigate the noise in the input scan {\small$\mathcal{L}_t$}. In this way, the {angular} distortion caused by the robot's ego-motion while capturing the data is corrected. 
Since each point has an associated acquisition timestamp, angular velocity measurements from the IMU can be integrated and then interpolated to compute the relative orientation w.r.t. the start of the scan.

\begin{figure} [t!]
    \centering
    \setlength{\belowcaptionskip}{-5pt}
    \includegraphics[width=0.9\linewidth]{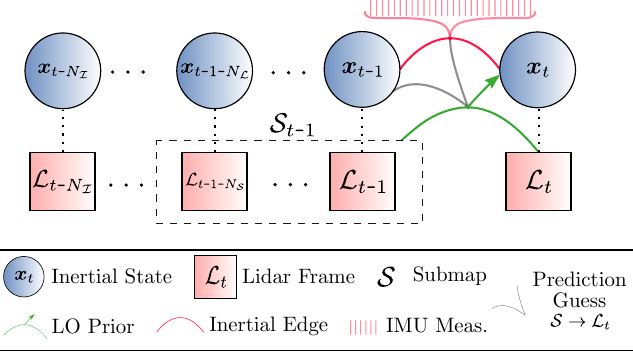}
    \caption{Scheme of how the different constraints are integrated into the graph-based LIO optimization framework.}
    \label{fig:lio-scheme}
\end{figure}

The de-skewed cloud is then aligned with the previous submap {\small$\mathcal{S}_{t-1}$} created by concatenating the {\small$ N_\mathcal{S}$} previous point clouds (see Figure~\ref{fig:lio-scheme}).
\emph{Scan-to-submap alignment} relies either on the efficient and GPU-accelerated implementation of voxelized generalized ICP (FastGICP)~\cite{koide2021fastgicp}, or the parallelized version of NDT (NDTOmp)~\cite{ndtomp}.
This process outputs LO estimates between frames (\textit{i.e.,}~the transformations {\small${}^{i\texttt{-}1}\tilde{\boldsymbol{T}}^{\text{LO}}_i$}), which can be introduced into the graph as prior edges to constrain the pose of the inertial state after computing \mbox{\small${}^i\tilde{\boldsymbol{T}}^{\text{LO}}= \left(\boldsymbol{T}_{i\texttt{-}1} \ {}^{i\texttt{-}1}\tilde{\boldsymbol{T}}^{\text{LO}}_i\right)^{\texttt{-}1}$}. The residuals for the {\small$i$}-th frame will be given by:
\begin{equation}
    \boldsymbol{e}_{i}^{\text{LO}} \triangleq \log\left(\boldsymbol{T}_i \ {}^i\tilde{\boldsymbol{T}}^{\text{LO}} \right)^\vee\in\mathbb{R}^6 \,,
\end{equation}
\noindent where the logarithm and vee operators are, respectively, {\small$\log(\cdot): SE(3)\mapsto \mathfrak{se}(3)$} and {\small$(\cdot)^\vee:\mathfrak{se}(3)\mapsto \mathbb{R}^6$}, being {\small$\mathfrak{se}(3)$} the Lie algebra of {\small$SE(3)$}.
The covariance of this measurement is inherited from the registration method~\cite{koide2021fastgicp}. Note that only unaccented variables will be optimization variables.

Simultaneously with the above,
IMU measurements between LiDAR frames are pre-integrated following~\cite{forster16}.
\mbox{\emph{Pre-integration}} provides an estimate of the robot's motion between successive frames, also allowing the scan-to-submap matching to be fed with a decent initial guess (see Figure~\ref{fig:lio-scheme}).
High-frequency odometry (at IMU rate) can be obtained by composing the LO with the relative motion predictions from the preintegrated measurement model.
IMU data is efficiently incorporated into the graph optimization framework by creating an inertial edge \cite{shan2020lio,campos21} between frames; this edge is actually composed of several different edges that correlate the poses and velocities between frames, and also model the bias variation over time (assumed to follow a Brownian motion pattern).
We use the method proposed in~\cite{forster16} to define the relative increments between frames and the preintegrated measurement model, \textit{i.e.,}~{\small$\Delta\tilde{\boldsymbol{R}}_{ij}$}, {\small$\Delta\tilde{\boldsymbol{v}}_{ij}$} and {\small$\Delta\tilde{\boldsymbol{p}}_{ij}$} (see~\cite{forster16} for details). This allows us to define their residual error as
{\small$\boldsymbol{e}^{\text{IMU}}\triangleq ( \boldsymbol{e}_{\Delta\boldsymbol{R}}, \boldsymbol{e}_{\Delta\boldsymbol{v}}, \boldsymbol{e}_{\Delta\boldsymbol{p}}, \boldsymbol{e}_{\Delta\boldsymbol{b}})^T\in\mathbb{R}^{15}$}, where:
\begin{subequations}
    \begin{alignat}{2}
        &\boldsymbol{e}_{\Delta\boldsymbol{R}_{i,j}} &&\triangleq \log(\Delta\tilde{\boldsymbol{R}}_{ij}^T \boldsymbol{R}_i^T \boldsymbol{R}_j)^\vee \,,\\
        &\boldsymbol{e}_{\Delta\boldsymbol{v}_{i,j}} &&\triangleq \boldsymbol{R}_i^T(\boldsymbol{v}_j\texttt{-}\boldsymbol{v}_i \texttt{-} \boldsymbol{g}\Delta t_{i,j}) \texttt{-}\Delta\tilde{\boldsymbol{v}}_{ij} \,, \\
        &\boldsymbol{e}_{\Delta\boldsymbol{p}_{i,j}} &&\triangleq \boldsymbol{R}_i^T(\boldsymbol{p}_j\texttt{-}\boldsymbol{p}_i\texttt{-}\boldsymbol{v}_i\Delta t_{i,j}\texttt{-}\frac{1}{2}\boldsymbol{g}\Delta t_{i,j}^2) \texttt{-} \Delta\tilde{\boldsymbol{p}}_{ij}\,,
    \end{alignat}
\end{subequations}
\noindent all of which live in {\small$\mathbb{R}^3$}, and {\small$\boldsymbol{e}_{\Delta\boldsymbol{b}_{i,j}}\triangleq \boldsymbol{b}_j-\boldsymbol{b}_i\in\mathbb{R}^6$}.
Note that the vee and logarithmic operators are now over {\small$SO(3)$}.

Finally, the graph containing the above constraints (see Figure~\ref{fig:lio-scheme}) is optimized using g2o~\cite{grisetti2011g2o}, solving the optimization:
\begin{equation}
   \mathcal{X}^\star = 
    \underset{\mathcal{X}}{\mathrm{argmin}}\left( \textstyle \sum\limits_{i\in\mathcal{I}_t} \| \boldsymbol{e}_{i}^{\text{LO}}\|^2 +  \sum\limits_{i,j\in\mathcal{I}_t} \| \boldsymbol{e}_{i,j}^{\text{IMU}}\|^2 \right)
\end{equation}

The success of registration algorithms depend on a good initial guess and sufficient overlap between the input clouds. The first issue has already been addressed, and the second is mitigated by the submaps, which in turn require some attention.
We use the scan-to-submap alignment to add new point clouds to the submap. Then, the latter is downsampled (\textit{e.g.,}~via uniform sampling) to prune redundant points.
To restrain complexity, the submap size is limited to include the last {\small$N_\mathcal{S}$} point clouds.
A side effect of this approach is high adaptability to different range sensors: we can handle densities from 16 to 128 beams and different patterns by changing {\small$N_\mathcal{S}$} 
By effectively configuring the submaps (\textit{e.g.,}~{\small$N_\mathcal{S}=1$} for 128-plane LiDARs), we keep the computational cost manageable while achieving the precision required for each application. 

%%%%%%%%%%%%%%%%%%%%%%%%%%%%%%%%%%%%%%%%%%%%%%%%%%%%%%%%%%%%%%%%%%%%%%%%%%%%%%%%
\subsection{3D Graph-based Localization}\label{S:4b}

\begin{figure} [t!]
    \centering
    \setlength{\belowcaptionskip}{-5pt}
    \includegraphics[width=.7\linewidth]{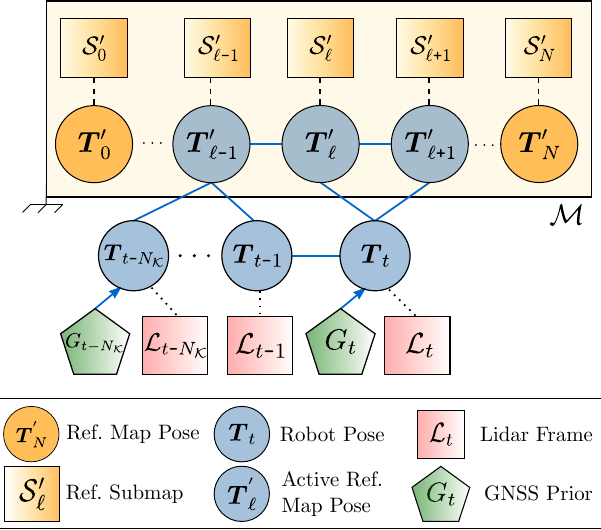}
    \caption{Overview of the proposed graph-based localization method using prior topological and metric information. Orange  elements form the \emph{reference graph}, and blue elements form the \emph{active graph}.}
    \label{fig:active_ref_graph}
\end{figure}

The global localization process is based on optimi\-zing the \emph{active graph}, which composed of a \emph{sliding graph} with the latest robot states ({\small$\boldsymbol{T}$}), and
regions of interest of the \emph{reference graph} (\textit{i.e.,}~certain reference vertices, {\small$\boldsymbol{T}'$}). See Figure~\ref{fig:active_ref_graph}.

The localization process starts with estimating the robot's initial pose within the prior map using GNSS measurements\footnote{In the absence of GNSS, this initial estimate can be given manually by the user, \textit{e.g.,}~using a ROS2 service.}. It then queries the prior topo-metric model to find the $k$-closest reference vertices (for the exemplary graph shown in Figure~\ref{fig:active_ref_graph}, at time {\small$t$}, they will be {\small$\boldsymbol{T}'_\ell$} and {\small$\boldsymbol{T}'_{\ell+1}$}).
Next, the current scan is aligned with the submaps associated with these vertices ({\small$\mathcal{S}_\ell',\mathcal{S}_{\ell+1}'$}).
The \emph{active graph} combines the constraints enforced by the results of these alignments with the relative motion data provided by the LIO and GNSS measurements (if available). 
Therefore, this new graph will contain vertices from both the prior reference graph and the online sliding graph. The latter encodes current and past robot states in a sliding window approach (\textit{i.e.,}~{\small$\mathcal{T}$}), being older nodes marginalized as they leave the window.
Figure~\ref{fig:gloc_general} displays the active (blue) and reference (orange) graphs on top of the prior point cloud map for a localization task. The online cloud is shown in red. 
Note how multiple reference vertices can be associated with a single vertex in the sliding graph (as is also shown in Figure~\ref{fig:active_ref_graph}).

The optimization of the active graph is governed by~\eqref{eq:optimization}, s.t.:
\begin{subequations}
  \allowdisplaybreaks
  \begin{alignat}{2}
    &\boldsymbol{e}_{i,j}^{\text{LIO}} &&\triangleq \log\left(\boldsymbol{T}_i \ {}^i\tilde{\boldsymbol{T}}^{\text{LIO}}_j \ (\boldsymbol{T}_j)^{\texttt{-}1} \right)^\vee\in\mathbb{R}^6 \,, \\
    &\boldsymbol{e}_{i}^{\mathcal{M}} &&\triangleq \textstyle \sum\limits_{k\in\mathcal{R}_i}\log\left(\boldsymbol{T}_i \ {}^i\tilde{\boldsymbol{T}}_k^{\mathcal{M}} \ (\boldsymbol{T}'_k)^{\texttt{-}1} \right)^\vee\in\mathbb{R}^6 \,, \\
    &\boldsymbol{e}_i^{\text{GNSS}} &&\triangleq \boldsymbol{p}_i\texttt{-}\tilde{\boldsymbol{p}}_i\in\mathbb{R}^3 \,, 
  \end{alignat}  
\end{subequations}
\noindent where {\small${}^i\tilde{\boldsymbol{T}}^{\text{LIO}}_j$} is the output from the previous subsection, {\small${}^i\tilde{\boldsymbol{T}}_k^{\mathcal{M}}$} are the registration results between the current scan and the prior map, and {\small$\boldsymbol{T}'_k\in\mathcal{G}$}.
Note that correspondences with the prior map are only evaluated for a subset of the reference vertices, \textit{i.e.,}~{\small$k\in\mathcal{R}_i\subset\mathcal{R}$}.
Finally, {\small$\tilde{\boldsymbol{p}}_i ={}^{\text{UTM}}\tilde{\boldsymbol{p}}_i\texttt{-} {}^{\text{UTM}}\tilde{\boldsymbol{p}}_{m}$}, with {\small${}^{\text{UTM}}\tilde{\boldsymbol{p}}_i$} the actual GNSS measurement and {\small$ {}^{\text{UTM}}\tilde{\boldsymbol{p}}_{m}$} the origin of the prior model (this origin will be associated with a Universal Transverse Mercator coordinate).
Relativizing GNSS allows to merge it with cloud-to-map matching and LIO constraints.

\begin{figure} [t!]    
    \centering
    \setlength{\belowcaptionskip}{-4pt}
    \includegraphics[width=0.9\linewidth]{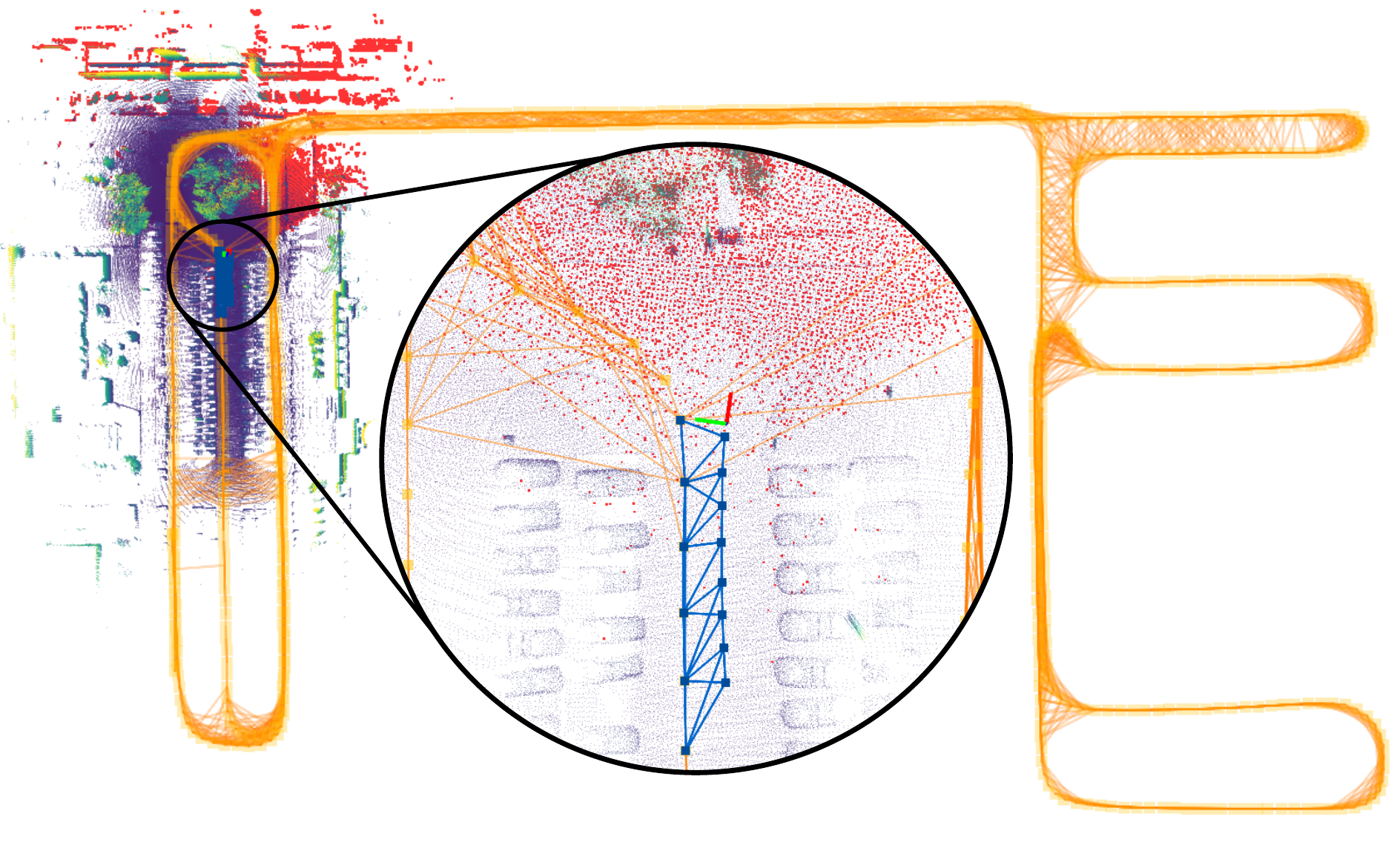}
    \caption{Visualization of dynamic loading. The prior graph is shown in orange, while the active graph is depicted in blue. For efficiency, only a subset of the geometric model is loaded, specifically the submaps corresponding to vertices within the robot's neighborhood.}
    \label{fig:gloc_qgl}
\end{figure}

On the one hand, LIO is used to connect sequential vertices of the sliding graph with the covariance depending on the alignment result (defined as the mean error between all matched points, weighted by the match percentage). In addition, it provides an initial guess for the cloud-to-map registration.
On the other hand, registrations between the current scan and the reference submaps are introduced as binary edges between vertices in the reference and sliding graphs, with information also depending on the alignment result.
Also, the transformation with the smallest covariance is used to compute the initial estimate for the optimization process (handled by g2o).
The previous constraints connect the two graphs and form the active graph (blue elements in Figure~\ref{fig:active_ref_graph}). 
Should the sliding and reference graphs become disconnected, either due to alignment failure or the lack of nearby reference vertices, the algorithm will rely on LIO and GNSS until re-localization occurs.
Finally, GNSS measurements are used as prior unary edges
when they are synchronized with the online cloud and if their uncertainty is below a certain threshold.

%%%%%%%%%%%%%%%%%%%%%%%%%%%%%%%%%%%%%%%%%%%%%%%%%%%%%%%%%%%%%%%%%%%%%%%%%%%%%%%%
\subsection{Dynamic Loading}\label{S:4c}
This module allows loading only nearby parts of the reference model, {\small$\mathcal{M}$}, as the robot moves; this is especially useful for large maps where maintaining (and visualizing) all submaps at high resolution is intractable.
Since both the reference and sliding graphs are georeferenced, it is possible to load the submaps near the current localization estimate in a sliding window fashion. 
To do this, as a new vertex is added to the sliding graph, we query the closest vertex in the reference graph and load its associated submap and those of its $k$-closest neighbors. 
While the number of reference submaps in memory is below a user-specified limit, the process is repeated after blacklisting the loaded submaps to avoid duplicates. In addition, the submaps that leave this window are removed from memory. 
This method allows us to control the total load by specifiying the maximum number of submaps (in practice, we load 15 to 30 submaps).
Figure~\ref{fig:gloc_qgl} illustrates the dynamic load of a portion of the environment.
Thus, we maintain a constant RAM memory usage during operation (\textit{e.g.,}~$\smallsim$50 MB for 30 submaps). In contrast, for some of the experiments in large environments that we present in the next section, loading the entire dense cloud would require up to 2 GB of memory.

%%%%%%%%%%%%%%%%%%%%%%%%%%%%%%%%%%%%%%%%%%%%%%%%%%%%%%%%%%%%%%%%%%%%%%%%%%%%%%%%
%%%%%%%%%%%%%%%%%%%%%%%%%%%%%%%%%%%%%%%%%%%%%%%%%%%%%%%%%%%%%%%%%%%%%%%%%%%%%%%%

\section{Experimental Results}\label{S:5}

\begin{table*}[t!]
    \scriptsize
    \centering
    \renewcommand{\arraystretch}{1.1}
    \begin{tabular}{l|c|c|c|c|c}
        \textbf{Dataset} & \textbf{Length (per sequence)} & \textbf{$\Delta$ Elevation} & \textbf{LiDAR} & \textbf{IMU} & \textbf{GNSS}\\
        \hline
        EULT (ref. \& eval.) & $\smallsim 5\phantom{.0}$ km & $\smallsim 50$ m & 2x Velodyne HDL-32E & XSens MTi-28A53G25 & Magellan ProFlex 500 \\ \hline
        NCE (ref. \& eval.) & $\smallsim 0.3$ km & $\smallsim \phantom{0}0$ m & \multicolumn{2}{c|}{Ouster OS0-128} & $-$ \\ \hline
        CUZ (ref.) & $\smallsim 2.7$ km & $\smallsim \phantom{0}1$ m & \multicolumn{2}{c|}{Ouster OS1-128} & \multirow{2}{*}{Ardusimple RTK2B}\\ \cline{1-5}
        CUZ (eval.) &$\smallsim 1.5$ km & $\smallsim \phantom{0}1$ m & \multicolumn{2}{c|}{Livox HAP} \\ \hline
        Bus & $\smallsim 8\phantom{.0}$ km & $\smallsim 70$ m & 2x RS Helios 32-5515 & \multicolumn{2}{c}{XSens MTI-680-G} \\ \hline
    \end{tabular}
    \caption{Summary of the sensor configurations and environmental characteristics in the experiments performed.}
    \label{tab:datasets}
\end{table*}

\begin{figure*} [t!]    
    \centering
    \setlength{\belowcaptionskip}{-10pt}
    \includegraphics[ width=0.82\linewidth]{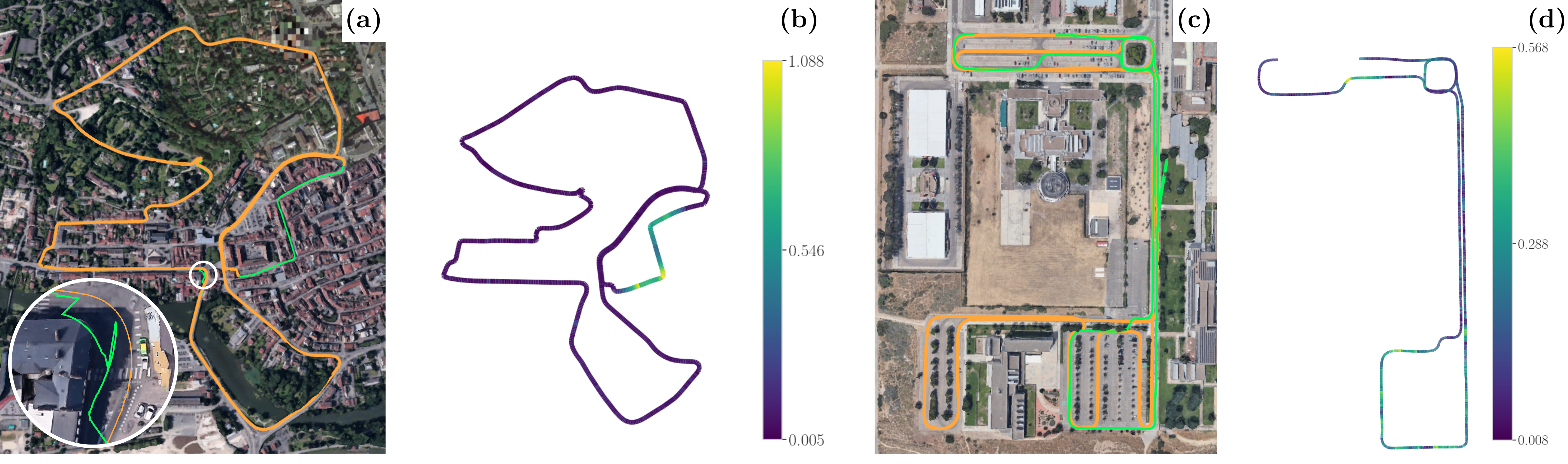}
    \caption{Evaluation results in EULT (a)-(b) and CUZ (c)-(d) datasets. (a), (c): georeferenced SLAM trajectories in the reference sequences (orange) and raw GNSS data in the evaluation sequences (green). (b), (d): Trajectories estimated by G-Loc in the evaluation sequences; the color map indicates 2D ATE (in meters, darker is smaller).}
    \label{fig:eult}
\end{figure*}

In this section, we present a thorough evaluation of G-Loc in different applications and under different sensor configu\-rations, demonstrating the versatility of our approach.
We use three real urban datasets: \emph{European Union Long-Term} (EULT)~\cite{yan20}, \emph{Newer College Extension} (NCE)~\cite{zhang21}, and our own dataset collected on the \emph{Campus of the University of Zaragoza} (CUZ).
In addition, we present the results of using G-Loc in a real-world automotive application, in the context of an autonomous driving bus.
The dataset experiments were performed on a laptop equipped with an Intel$^\text{{\tiny{\textregistered}}}$ Core$^\text{{\tiny{TM}}}$ i7-1165G7 CPU and an Nvidia GeForce$^\text{{\tiny{\textregistered}}}$ RTX$^\text{{\tiny{\textregistered}}}$ 3080 GPU, while the bus was equipped with an Intel$^\text{{\tiny{\textregistered}}}$ Core$^\text{{\tiny{TM}}}$ i9-12900TE and an Nvidia RTX$^\text{{\tiny{\textregistered}}}$ A2000 (note that many other autonomous driving processes were running concurrently on this hardware). 
Unless otherwise noted,~\cite{koide2021fastgicp} was used for registration.
Table~\ref{tab:datasets} contains a short summary of the experiments and the sensors installed in the mobile platforms. 
The error metrics presented in this section were obtained using {\small\texttt{evo}}~\cite{evo}.

%%%%%%%%%%%%%%%%%%%%%%%%%%%%%%%%%%%%%%%%%%%%%%%%%%%%%%%%%%%%%%%%%%%%%%%%%%%%%%%%
\subsection{Dataset Evaluation}

The \emph{EULT} dataset contains several sequences recorded with a car in a downtown. They all follow a similar trajectory under different conditions (\textit{e.g.,}~traffic, weather); this is particularly interesting for testing map-based localization, since we need different but overlapping reference and evaluation trajectories. Sequence and sensor details are shown in Table~\ref{tab:datasets}.

To evaluate the performance of G-Loc, we first generated a georeferenced model of the environment using one of the available sequences, which we will refer to as the \emph{reference sequence}\footnote{\label{fn:ids}Ref. seq. ID: \emph{2018-07-19}. Eval. seq. ID: \emph{2018-07-16} and \emph{2018-07-17}.}.
To obtain the required prior model with topological (a pose-graph) and metric (point cloud submaps) information, we used~\cite{montano24}, a SLAM system with a similar architecture to G-Loc but replacing the localization module with a mapping module for robust loop closing and global optimization.
To evaluate the performance in mapless regions, we removed an entire road from the prior model ({$\smallsim0.4$ km); the SLAM-estimated trajectory is shown in orange in Figure~\ref{fig:eult}a.
The built model was fed into G-Loc while running two different \emph{evaluation sequences}\footref{fn:ids}. Figure~\ref{fig:eult}a also depicts the raw GNSS data (green) for one of them to facilitate comparison with the reference trajectory. The middle right section contains the cropped region, where there is GNSS signal but no prior information.
When the robot traverses a mapped region, the prior model is exploited to improve localization and correct for potential drift in LIO. Conversely, in unmapped regions, the sliding graph disconnects from the reference graph, accumulating a larger error that is eventually corrected after re-localization. 
Global measurements in this large mapless region kept the absolute error bounded, enabling the system to re-localize with the map as it traversed previously mapped areas.
Figure~\ref{fig:eult}b contains the 2D Absolute Trajectory Error (ATE) for translation along the trajectory for one of the sequences\footnote{
Due to the lack of ground-truth in this dataset and the noise levels of GNSS in certain regions, errors were calculated w.r.t. SLAM. This algorithm fuses GNSS with LIO to mitigate the effects of noisy measurements and GNSS-denied areas (see zoomed region of Figure~\ref{fig:eult}a). The error between GNSS and SLAM is negligible when the former is reliable.}, showing that the highest errors occur on the unmapped road and that even under these conditions they are within reasonable limits (see colormap).
Table~\ref{tab:results} contains the numerical results (mean and standard deviation) for the cropped sequences, namely Root Mean Squared Error (RMSE) of 3D ATE for translation and rotation, and lateral and longitudinal error. 
Furthermore, we report the results of using a complete prior map (\textit{i.e.,}~uncropped); in this case, the errors were similar but exhibited a smaller deviation. 
We also evaluated the use of NDTOmp, which is of particular interest for resource-constrained platforms. In this case, the overhead slightly increased the localization errors.

Finally, to benchmark G-Loc, we conducted the same experiments using Autoware~\cite{kato15}, a popular localization system for autonomous vehicles. 
For a fair comparison, we fed velocity measurements from our LIO module and used NDTOmp.
Autoware failed to localize in the mapless region, leading to an unrecoverable state. Using the complete prior model, the translation errors were comparable to G-Loc but the rotation error was nearly doubled (see Table~\ref{tab:results}). In addition, Autoware required the use of an Intel{$^\text{{\tiny{\textregistered}}}$ Core$^\text{{\tiny{TM}}}$ i9 CPU to work properly.

\begin{table*}[t!]
    \scriptsize
    \centering
    \setlength{\belowcaptionskip}{-4pt}
    \renewcommand{\arraystretch}{1.1}
    \begin{tabular}{c|l||l|l|l|l|l}
        \textbf{Dataset} & \textbf{Method} & \textbf{ATE (trans., m)} & \textbf{Long. Error (m)} & \textbf{Lat. Error (m)} & \textbf{ATE (rot., deg)} & \textbf{Time per frame (ms)}\\ \hline %\hline
        \multirow{3}{*}{\makecell{EULT\\(cropped)}} & G-Loc (FastGICP) &$\phantom{{}^\dagger}$\textbf{0.209 {\tiny (0.395)}} &$\phantom{{}^\dagger}${0.087 {\tiny (0.280)}} &$\phantom{{}^\dagger}${0.048} {\tiny (0.228)} &$\phantom{{}^\dagger}$\textbf{0.423 {\tiny (0.445)}} &$\phantom{{}^\dagger}$\textbf{25.13 {\tiny (7.82)}} \\ %\cline{2-7}
        & G-Loc (NDTOmp) &$\phantom{{}^\dagger}${0.240} {\tiny (0.300)} & $\phantom{{}^\dagger}$\textbf{0.086} {\tiny (0.304)} &$\phantom{{}^\dagger}$\textbf{0.021 {\tiny (0.208)}} &$\phantom{{}^\dagger}${1.008} {\tiny (1.160)} &$\phantom{{}^\dagger}${38.01} {\tiny (14.86)} \\ %\cline{2-7}
        & Autoware &$\phantom{{}^\dagger}\times$ &$\phantom{{}^\dagger}\times$ &$\phantom{{}^\dagger}\times$ &$\phantom{{}^\dagger}\times$ & $\phantom{{}^\ddagger}\times$\\ \hline %\hline
        \multirow{3}{*}{\makecell{EULT\\(complete)}} & G-Loc (FastGICP) &$\phantom{{}^\dagger}$\textbf{0.194 {\tiny (0.201)}} &$\phantom{{}^\dagger}$\textbf{0.101 {\tiny (0.240)}} &$\phantom{{}^\dagger}$0.014 {\tiny (0.197)} &$\phantom{{}^\dagger}$\textbf{0.440 {\tiny (0.576)}} & $\phantom{{}^\ddagger}$\textbf{26.47 {\tiny (4.31)}}\\ %\cline{2-7}
        & G-Loc (NDTOmp) &$\phantom{{}^\dagger}$0.218 {\tiny (0.196)} &$\phantom{{}^\dagger}${0.119} {\tiny (0.245)} &$\phantom{{}^\dagger}${0.009} {\tiny (0.203)} &$\phantom{{}^\dagger}$0.953 {\tiny (0.836)} & $\phantom{{}^\ddagger}$40.13 {\tiny (12.99)}\\ %\cline{2-7}
        & Autoware &$\phantom{{}^\dagger}${0.212} {\tiny (0.194)} &$\phantom{{}^\dagger}$0.123 {\tiny (0.232)} &$\phantom{{}^\dagger}$\textbf{0.006 {\tiny (0.203)}} &$\phantom{{}^\dagger}${0.784} {\tiny (1.380)} & ${}^\ddagger${37.31} {\tiny (14.52)}\\ \hline %\hline
        \multirow{3}{*}{CUZ} & G-Loc (FastGICP) &$\phantom{{}^\dagger}$\textbf{0.154 {\tiny (0.102)}} &$\phantom{{}^\dagger}$\textbf{0.071 {\tiny (0.132)}} &$\phantom{{}^\dagger}${0.019} {\tiny (0.126)} &$\phantom{{}^\dagger}$\textbf{0.473 {\tiny (0.537)}} & $\phantom{{}^\ddagger}$\textbf{34.62 {\tiny (9.14)}}\\ %\cline{2-7}
        & G-Loc (NDTOmp) &$\phantom{{}^\dagger}${0.214} {\tiny (0.151)} & $\phantom{{}^\dagger}${0.131} {\tiny (0.179)} &$\phantom{{}^\dagger}$\textbf{0.014 {\tiny (0.158)}} &$\phantom{{}^\dagger}${0.650} {\tiny (0.878)} & $\phantom{{}^\ddagger}$72.15 {\tiny (31.54)}\\ %\cline{2-7}
        & Autoware &${}^\dagger$0.308 {\tiny (1.251)} &${}^\dagger$0.400 {\tiny (0.655)} &${}^\dagger$0.086 {\tiny (1.353)} &${}^\dagger$2.146 {\tiny (0.850)} & ${}^\ddagger${59.74} {\tiny (71.47)} \\ %\hline
    \end{tabular}
    \caption{Localization errors (mean and std. dev.) and time consumed per frame for G-Loc (FastGICP and NDTOmp) and Autoware in EULT and CUZ datasets. Best results are bold, and $\times$ indicates an unrecoverable failure in localization.}
    \label{tab:results}
\end{table*}

The \emph{CUZ} dataset was captured using a sensor-equipped car. We recorded two sequences, which partially overlap, but with two different LiDAR sensors. See Table~\ref{tab:datasets}.
To evaluate this dataset, we followed a similar procedure as in EULT.
First, we built the prior model with a reference sequence using a high-performance sensor with $360\degree$ Field of View (FoV) and $128$ planes. Then, we evaluated G-Loc with an evaluation sequence, exploiting the prior topo-metric know\-ledge and using a LiDAR with $120\degree$ FoV and a completely different sensing pattern. 
Figure~\ref{fig:eult}c displays the trajectory of the sequence (orange) and the raw GNSS data for the evaluation one (green). Both trajectories overlap for most of the route, and when they do not, they are close enough so that the prior map can be used (unlike in EULT). ATE error along the trajectory is shown in Figure~\ref{fig:eult}d and numerical results are reported in the bottom of Table~\ref{tab:results}.
As in EULT, the use of FastGICP gave the best results, except for the lateral error (however, these lower mean values are accompanied by higher variance). 
This experiment demonstrates the adaptability of our method to the use of different sensors for building the prior map and localization. The submaps played a fundamental role here, mitigating the effects of the different sensing patterns and feeding the registration algorithms with sufficient overlapping regions.
In contrast, Autoware accumulated much higher errors and had difficulty finding correspondences in many regions (which explains the large deviations), eventually leading to an unrecoverable state. The results reported in Table~\ref{tab:results} for this experiment (marked with ${}^\dagger$) are only before this divergence.

The \emph{NCE} dataset contains several sequences recorded using a handheld device in Oxford. 
This dataset allows us to test G-Loc under different conditions than the previous ones.
It includes aggressive motion, where IMU integration is crucial for undistorting the point cloud and predicting the motion between successive LiDAR frames.
We have used \emph{Quad}, \emph{Math}, and \emph{Underground} collections of NCE, each of which contains three sequences of increasing difficulty and slightly different trajectories.
First, we used~\cite{montano24} to build a prior model with the \emph{easy} sequences and then ran our system in the \emph{medium} and \emph{hard} ones.
Figure~\ref{fig:gloc_general} shows the resulting prior model for the \emph{Math} collection. We compared the performance of G-Loc with two SLAM (LIO-SAM~\cite{shan2020lio}, NV-LIO~\cite{chung24}) and two map-based localization methods (LiDAR-Loc~\cite{rsasaki23}, HDL-Loc~\cite{koide17}). 
For a fair comparison, all localization methods used the same prior map. 
Table~\ref{tab:nce_benchmark} contains ATE RMSE, with the values for the SLAM methods extracted from~\cite{chung24}. 
The results showcase the robustness and accuracy of our system, which outperformed the localization methods and succeeded in every sequence. The failures of 
\cite{rsasaki23,koide17} were caused by the lack of prior maps in certain regions (\emph{Quad-h}) and aggressive motion (\emph{Math-h}). 
G-Loc maintained accurate localization in these regions thanks to LIO, allowing for re-localization.
G-Loc also outperformed the SLAM methods in most sequences, achieving an average error of 8~cm (note that the
localization errors ultimately depend on the error of the prior model, see~\cite{montano24} for detailed values).

\begingroup
\setlength{\tabcolsep}{4pt} % Default value: 6pt
\begin{table}[t!]
    \centering
    \scriptsize
    \begin{tabular}{r|cc|cc|cc|c}
        \textbf{Method / Seq.} &Quad-m &Quad-h &Math-m &Math-h &Und-m &Und-h & Avg.\\ \hline 
        LIO-SAM~\cite{shan2020lio} &\textbf{0.067} & \underline{0.13} &0.13 & \textbf{0.085} & \textbf{0.065} & 0.42 & 0.15\\
        NV-LIO~\cite{chung24} & \underline{0.076}  & 0.18 & \underline{0.095} & 0.094 & \underline{0.072} & \underline{0.078} & \underline{0.10} \\
        LiDAR-Loc~\cite{rsasaki23} &0.11 &$\times$ &0.10 &$\times$ &0.11 &$\times$ &0.11 \\
        HDL-Loc~\cite{koide17} &0.10 &$\times$ &0.11 &$\times$ &0.11 &0.12 &0.11 \\ \hline
        G-Loc (\emph{ours}) &\textbf{0.067} & \textbf{0.087} & \textbf{0.090} & \underline{0.093} & 0.074 & \textbf{0.075} &\textbf{0.081}    
    \end{tabular}
    \caption{ATE RMSE (m) in NCE dataset. Best results are bold and second best are underlined. $\times$ indicates an unrecoverable failure.}
    \label{tab:nce_benchmark}
\end{table}
\endgroup

%%%%%%%%%%%%%%%%%%%%%%%%%%%%%%%%%%%%%%%%%%%%%%%%%%%%%%%%%%%%%%%%%%%%%%%%%%%%%%%%
\subsection{On the Time Consumption}

While the above section aimed to demonstrate the robustness and precision of our approach, we now seek to demonstrate its real-time performance. 
The processing times per frame consumed by each algorithm in the conducted experiments are listed in Table~\ref{tab:results}. G-Loc required, on average, $25$ ms in EULT and $34$ ms in CUZ ---this is much faster than the typical LiDAR frame rate ($10$ Hz) and therefore meets real-time requirements. In both cases, LIO accounts for $60\%$ of that time, and the localization module for the rest (visualization and dynamic load are excluded as they operate in the background). 
G-Loc with NDTOmp took considerably longer, but the frame rate was generally not exceeded.
We also report the time consumed by Autoware (marked with ${}^\ddagger$, as it used a better CPU), which is similar to G-Loc using NDTOmp.

Finally, we compared the use of G-Loc to running a full SLAM system to reveal the benefits of exploiting the known prior topo-metric map instead of building a new map. We used the EULT dataset because it is the largest and therefore where a new map would be more costly to generate and maintain.
The time required to process a frame is similar to G-Loc (the LIO update), but the graph optimization and loop closing took $190.4 \pm 101.4$ ms.
These demanding processes slowed down the entire pipeline and caused interruptions in the localization.

%%%%%%%%%%%%%%%%%%%%%%%%%%%%%%%%%%%%%%%%%%%%%%%%%%%%%%%%%%%%%%%%%%%%%%%%%%%%%%%%
\subsection{Real-world Deployment}

G-Loc has been deployed on an autonomous bus (see Figure~\ref{fig:digizity}a) on an urban bus line in the city of Zaragoza (Arag\'on, Spain) as part of the pioneering Spanish project on urban driving, DIGIZITY. 
The sensors installed on the vehicle and the route specifics appear in Table \ref{tab:datasets}. 
Our system requires coordination with other navigation systems to provide a robust and reliable localization, ensuring safe navigation even in areas with poor GNSS coverage.
The prior model of the entire route is shown in Figure~\ref{fig:digizity}b. In addition, localization starts as soon as the bus is turned on. This means that the localization has to work without a prior model from the departure point until the route is reached ($\smallsim 2$ km). 
The system localizes in the unmapped region using LIO and GNSS in a sliding-graph optimization fashion until re-localization with the prior model occurs. 
Figure~\ref{fig:digizity}c depicts the trajectory estimated by G-Loc for a single trip, and Figure~\ref{fig:digizity}d shows the improvement of these estimates (blue) over the raw GNSS data (green) in one of the many low coverage areas.
During the deployment, the bus operated seamlessly for about $100$ hours over a span of $20$ days, covering about $800$ km. Notably, there were no localization failures and the errors remained within bounds comparable to previous results (\textit{i.e.,} at the cm level), further showcasing the robustness and accuracy of G-Loc.

%%%%%%%%%%%%%%%%%%%%%%%%%%%%%%%%%%%%%%%%%%%%%%%%%%%%%%%%%%%%%%%%%%%%%%%%%%%%%%%%
%%%%%%%%%%%%%%%%%%%%%%%%%%%%%%%%%%%%%%%%%%%%%%%%%%%%%%%%%%%%%%%%%%%%%%%%%%%%%%%%
\section{Conclusion}\label{S:6}

In this work we have presented G-Loc, a localization system that seamlessly integrates GNSS data, LIO and cloud-to-map registration into a graph optimization framework.
We have proposed a novel method for map-based localization that leverages the results of a previous graph-SLAM process, utilizing both the geometric map and the underlying pose-graph representation. 
We have tested G-Loc in several datasets and in a real-world automotive application, demonstrating its accuracy, robustness, and adaptability to different sensor configurations and environmental conditions (\textit{e.g.,}~the lack of prior knowledge in some parts of the environment).
Long-term autonomy requires the sporadic execution of mapping to update the robot's knowledge of the environment. 
This work sets the stage for developing a long-term approach where the prior model is updated and refined with online information.

\begin{figure}[t!]
    \centering
    \setlength{\belowcaptionskip}{-2pt}
    \includegraphics[width=.92\linewidth]{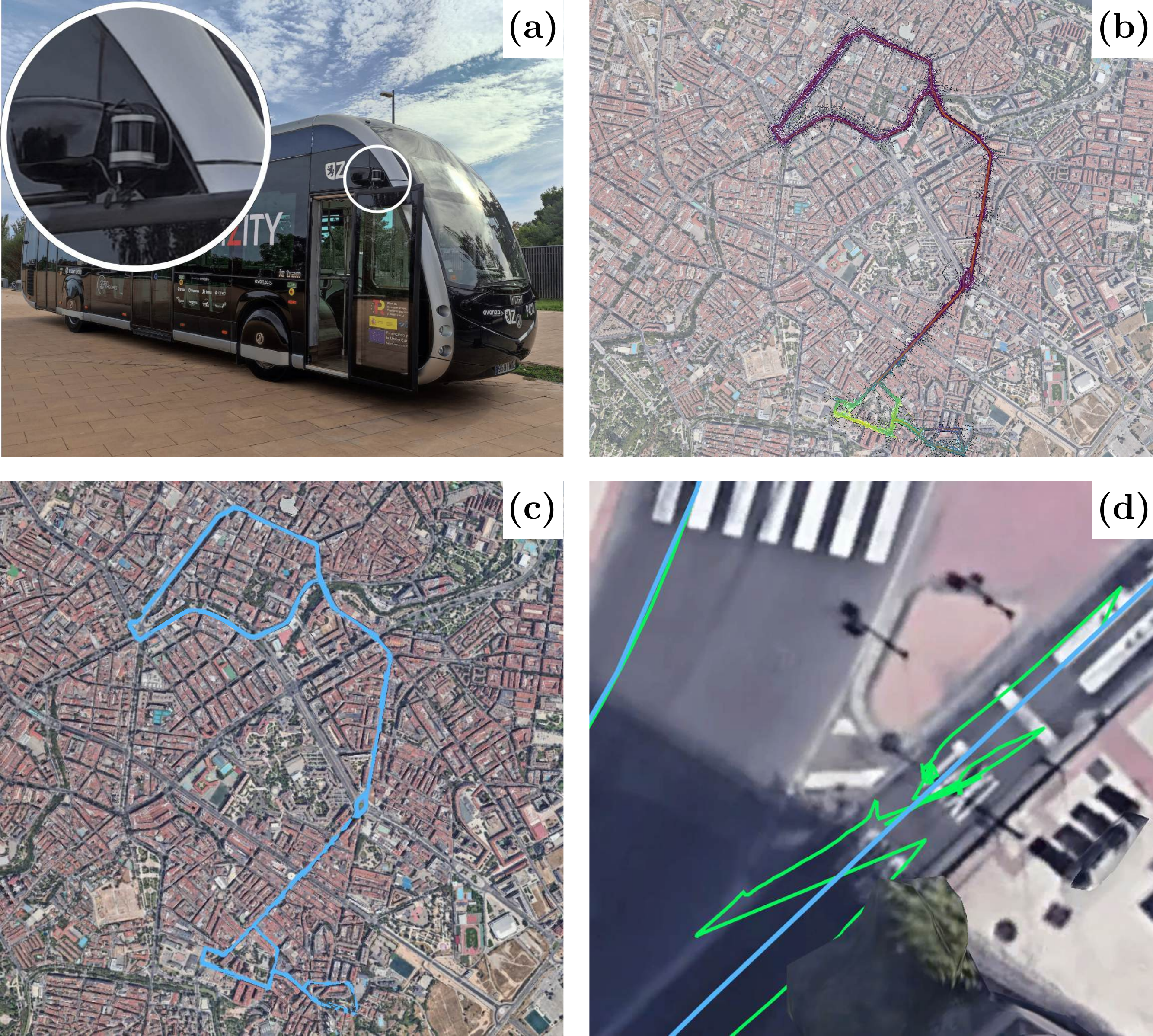}
    \caption{Results of the real-world experiment. (a) Platform and sensor setup. (b) Prior topo-metric model (colored by height, darker is lower). (c) Complete trajectory estimated by G-Loc. (d) Comparison between the raw GNSS (green) and the localization estimates (blue).}
    \label{fig:digizity}
\end{figure}

%%%%%%%%%%%%%%%%%%%%%%%%%%%%%%%%%%%%%%%%%%%%%%%%%%%%%%%%%%%%%%%%%%%%%%%%%%%%%%%%
%%%%%%%%%%%%%%%%%%%%%%%%%%%%%%%%%%%%%%%%%%%%%%%%%%%%%%%%%%%%%%%%%%%%%%%%%%%%%%%%
\ifCLASSOPTIONcaptionsoff
  \newpage
\fi

\bibliographystyle{IEEEtran}
\bibliography{bibliography}

\end{document}